\documentclass{article}

\usepackage{PRIMEarxiv}

\usepackage[utf8]{inputenc} 
\usepackage[T1]{fontenc}    
\usepackage{hyperref}       
\usepackage{url}            
\usepackage{booktabs}       
\usepackage{amsfonts}       
\usepackage{nicefrac}       
\usepackage{microtype}      
\usepackage{lipsum}
\usepackage{fancyhdr}       
\usepackage{graphicx}       
\usepackage{booktabs}
\graphicspath{{media/}}     

\title{Multi-modal reward for visual relationships-based image captioning
}

\author{
  Ali Abedi \\\
  University of Isfahan \\\
   \And
  Hossein Karshenas \\
  University of Isfahan \\
    \And
  Peyman Adibi \\
  University of Isfahan \\
    \And
}

\begin{document}
\maketitle

\begin{abstract}
Deep neural networks have achieved promising results in automatic image captioning due to their effective representation learning and context-based content generation capabilities. As a prominent type of deep features used in many of the recent image captioning methods, the well-known bottom-up features provide a detailed representation of different objects of the image in comparison with the feature maps directly extracted from the raw image. However, the lack of high-level semantic information about the relationships between these objects is an important drawback of bottom-up features, despite their expensive and resource-demanding extraction procedure. To take advantage of visual relationships in caption generation, this paper proposes a deep neural network architecture for image captioning based on fusing the visual relationships information extracted from an image’s scene graph with the spatial feature maps of the image. A multi-modal reward function is then introduced for deep reinforcement learning of the proposed network using a combination of language and vision similarities in a common embedding space. The results of extensive experimentation on the MSCOCO dataset show the effectiveness of using visual relationships in the proposed captioning method. Moreover, the results clearly indicate that the proposed multi-modal reward in deep reinforcement learning leads to better model optimization, outperforming several state-of-the-art image captioning algorithms, while using light and easy to extract image features. A detailed experimental study of the components constituting the proposed method is also presented.
\end{abstract}


\section{Introduction}
\begin{figure}
  \centering
  \includegraphics{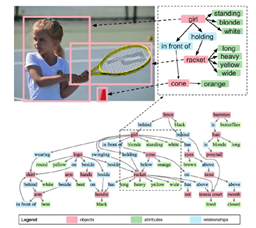}
  \caption{An example of a scene graph (bottom) and a grounding (top) [33].}
  \label{fig:fig1}
\end{figure}

Image captioning is the task of automatically generating a caption to describe the image content. This task is challenging because it should take an image as input and describe it using natural language sentences. Therefore, it combines machine vision to extract the image’s features and natural language processing to convert these features to human language [1], [2]. This task has several applications such as image indexing [3], helping visually impaired people [4], and content-based image retrieval [5]. Recently, deep neural networks have shown remarkable results in image captioning [6]–[10]. Among several existing architectures of deep neural networks for image captioning, the encoder-decoder framework has been widely used in the state-of-the-art models, because of the multi-modal nature of this task [6], [7], [11]–[13]. In the encoder-decoder framework for image captioning, the encoder usually consists of a convolutional neural network (CNN)-based architecture to extract the feature sets of an image. These features are then fed into the decoder, usually a recurrent neural network (RNN) (e.g., long short-term memories - LSTM) as a language model, to generate the final captions. \\
In encoder-decoder architecture, the decoder generates each caption’s tokens considering the previously generated tokens and the image features from the encoder. Two issues raise in this approach: (1) The model is trained using maximum likelihood estimation, causing the exposure bias problem [14], which makes the generated captions look like ground-truth captions. In other words, in the training process, the model is exposed to the training data, while in the inference time, the model is exposed to the model's output tokens in each time step. As a result, an error in generating tokens may cause a completely bad result. (2) Image captioning evaluation metrics at the inference time are non-differentiable, so they cannot be used as loss functions. For a sequence model, it is better to be optimized directly on test-time metrics. To accomplish this, recent image captioning methods use deep reinforcement learning (DRL) to optimize the network to tackle these problems [10], [15], [16]. A reinforcement learning agent chooses an action in an environment, receives a reward for that action, and changes its state. The better the action, the greater the reward. The agent's goal is to choose actions that maximize the expectation of having greater long-term rewards. DRL-based image captioning methods sample the next token from a deep model based on the reward of each state [1]. \\
Policy gradient methods for DRL have shown remarkable ability to train deep neural networks directly on non-differentiable metrics. These methods optimize the action generating policy model using a gradient descent method [1], [17]. REINFORCE [18] is a policy-gradient-based algorithm which introduces a baseline in the gradient estimation formula to reduce the variance of the gradient estimate. Self-critical sequence training (SCST) [18] is a version of REINFORCE algorithm which directly uses CIDEr captioning metric [18] as reward, normalized with the inference time output as baseline. This metric, which is especially designed for image captioning task, compares the generated caption with the ground truth caption given for an image. Since the contents of an image can be described using different terminologies, when merely using CIDEr as the reward function the captioning tend to memorize the true captions, leading to overfitting. In other words, such a linguistic-only reward function cannot consider the visual information of the input image, and ignores the correspondence between image and the generated caption. \\
Many of the recent image captioning models [7], [13], [14]–[16] employ the so called bottom-up features for considering the visual information of images. To obtain these features usually a pre-trained faster region-based CNN [22] model is used first to extract image objects bounding boxes. Then, these bounding boxes are fed into a CNN (e.g., Resnet [23]) to compute the high-level representative features. Also, several studies tend to utilize semantic information as the image features to improve the image captioning results [24]–[28]. While bottom-up features provide more extensive semantic information about the entities in the image than simpler feature maps directly extracted from the raw image, they do not consider relationships between image objects which are vital in generating captions. From a semantic point of view mentioning the relationships between objects would enrich the generated captions. What is more, extracting the images bounding boxes and then extracting each bounding box’s feature maps are very time-consuming, requiring powerful resources, and resulting in large feature maps which cause these features to be usually pre-extracted. \\
High-level visual relationships in an image can provide vital complementary semantic information for image captioning systems when properly fused with the feature maps extracted from the images. One of the best solutions to represent semantic visual relationships in an image is to use the scene graph [29]. As shown in Figure \ref{fig:fig1} scene graphs are structures that, instead of merely containing the images objects, also contain object relationships and (sometimes) the objects attributes [30], [31]. Given a set of object classes $C$ and a set of relationship classes $R$, a scene graph $G$ is defined as $G=(O,E)$, where $O=\{o_1,o_2,…,o_n \} \subseteq C$ is the set of image objects which are graph nodes and $E \subseteq O \times R \times O$ is the set of edges and represent the relationships between two objects [32]. \\
There are several entities and relationships in an image, and it is important to consider only the most salient objects and relationships. Inspired by machine translation [34], [35], the attention mechanism is used in encoder-decoder frameworks for image captioning. The attention mechanism introduces weight vectors used in each time step to dynamically focus on different parts of the input image containing an object during the caption generation process. Although attention mechanism has achieved remarkable results in various fields, in traditional attention blocks the attention result may not be what the decoder expects, thus misleading the decoder in result generation because of a mistaken belief. This problem could happen in two different situations: (1) there are errors in the attention module output or (2) there is no valuable information in the candidate vectors [20]. The first type of errors can be reduced by better attention modeling, but cannot be completely removed. To tackle the latter issue, attention on attention (AoA) [20] has been proposed, which extends the conventional attention mechanism by adding another attention step. Such an AoA block can be employed to identify important visual relationships that can influence the final generated caption. However, proper fusion of these relationships with other visual and linguistic information in the image captioning model, remains a challenge which is addressed in the present study. \\
This paper proposes a deep neural model for image captioning to overcome the aforementioned drawbacks and challenge. First, visual relationship triplets are introduced in this study as a three-word sentence $T=<o_1, r, o_2>$, extracted from the scene graph of an image, where $o_1$ and $o_2$ are two image objects, and r is the relationship between them. Then, a deep neural architecture based on AoA block is proposed to fuse these visual relationship features with feature maps extracted from the image using a fast and light CNN (e.g., Resnet). This allows the captioning system to exploit both spatial and semantic features simultaneously, and makes end-to-end training of the system possible. Finally, a multi-modal reward function is proposed for DRL in a two-phase supervised-reinforced training process of the captioning system. This reward function consists of CIDEr, as the language reward, and an image-caption similarity as vision reward. The language reward measures the similarity between the generated and ground truth captions. The vision reward measures the similarity between the generated caption and the image contents, obtained from an embedding network for mapping both caption and image into a common latent space. The experimental results on the standard MSCOCO image captioning dataset [36] show an improved or comparable performance comparing with several state-of-the-art methods in fair conditions. Detailed experimental analysis reveal the effectiveness of the visual relationships information on improving the performance of the captioning system based on common evaluation metrics. Moreover, the results indicate that using the proposed multi-modal reward for DRL with SCST algorithm outperforms the original version of this algorithm. \\
In summary, the main contributions of this study are as follows:
\begin{itemize}
	\item Scene graph-based visual relationship triplets are introduced as high-level semantic features to improve the image captioning results.
	\item An AoA-based deep neural architecture is proposed for fusion of visual relationships and spatial features extracted from images in the decoder of the captioning system.
	\item A new multi-modal reward function is proposed for optimizing image captioning systems with DRL. This reward combines language and visual similarity with the ground truth captions and images, respectively.
 \end{itemize}
The rest of the paper is organized as follows. In Section 2 the related works are reviewed. The proposed method will be explained in Section 3. The dataset, experimental settings, and the achieved experimental results are presented in Section 4. Finally, several concluding remarks and possible future works are given in Section 5.

\section{RELATED WORK}
\label{sec:relatedWork}

\subsection{Image Captioning}
The existing image captioning works are divided into three main categories. First, there are template-based image captioning systems that use fixed templates with some blank tokens. These tokens should be generated using image features, which are usually hand-crafted [1], [37]–[40]. Although these models are easy to train and use, they cannot generate captions with different lengths. Moreover, because the blank spots are limited, they cannot mention all the essential objects and relations of the image, and they are completely limited to their templates. \\
Second category is retrieval-based image captioning. These methods create a caption pool using ground-truth captions in training data. To generate a caption for a specific image, they collect the most similar images to query images from the caption pool. The final caption is generated by selecting the most suitable caption or combining the retrieved captions [41]–[45]. Because the captions in the caption pool are human-generated, the captions generated using this method are syntactically correct. However, there is a high probability of generating irrelevant captions if the query image has some objects or relations that do not appear in the training dataset. The inherent problems and limitations in the first two categories lead the machine vision scientific community to the third category of image caption generation, which is the most general way to generate a caption for an image. \\
The third group contains deep learning methods which have shown significant performance in complicated tasks like image captioning. The state-of-the-art image captioning methods, use an encoder-decoder framework to build the image captioning systems [6], [7], [13], [20], [24], [26], [28]. Vinyals et al. [6] used a deep CNN to extract image features and fed these features into an RNN, and trained the language model conditioned on image features. Jia et al. [28] guided the model by adding semantic information extracted from the image to each unit of an LSTM. Mao et al. [46] proposed a multi-modal Recurrent Neural Networks (m-RNN) model. First, it uses a deep CNN to encode the image and an RNN language model to encode previously generated words. Then a multi-modal model combines visual and textual information to predict the next word. Chen et al. [47] proposed a bi-directional representation capable of generating both novel descriptions from images and visual representations from descriptions.

\subsection{Attention Mechanism}
\begin{figure}
  \centering
  \includegraphics{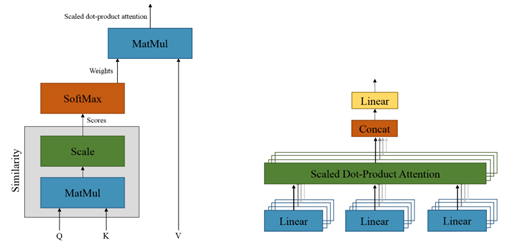}
  \caption{Scaled dot-product attention (left) and multi-head attention (right).}
  \label{fig:fig2}
\end{figure}

There is lots of information in an image, and it is unnecessary to describe all the entities. Attention mechanism [48] is one of the concepts that originated from machine translation [34] and is widely used in image captioning tasks as well [19], [26], [49]–[54]. The attention mechanism is used to guide encoder-decoder-based image captioning models. The block diagram of the attention and multi-head attention modules [48] have been illustrated in Figure \ref{fig:fig2}. The attention module takes queries (Q), keys (K), and values (V) as input and generates weight vectors. First, it measures the similarities between Q and K and then feeds the scores into a softmax function to calculate the weights. Once the weights are calculated, they are applied to V. In neural networks, using an attention mechanism allows the network to have the ability to focus on a subset of input features and select the more salient parts of the image’s features by calculating an importance score for each feature vector and then applying these scores to those feature vectors [7]. The Q, K and V are linear transformations of the input image features. Instead of computing attention once, multi-head attention runs attention function multiple times in parallel. This allows the model to attend to different parts of the input sequence simultaneously [48].\\
The calculation process of multi-head attention has been shown in Equations \ref{equation:eq1}, \ref{equation:eq2}, \ref{equation:eq3}. Each head calculated by Equation \ref{equation:eq3}. Equation \ref{equation:eq4} shows the softmax formula where $d$ is the dimension of $Z_i$ and $F$ is the number of $Z_i$ elements.

\begin{equation}
    f_{mh-att} (Q, K, V)=Concat(head_1, head_2, …, head_H)
    \label{equation:eq1}
\end{equation}
\begin{equation}
    head_i=f_{dot-att} (Q_i, K_i, V_i)
    \label{equation:eq2}
\end{equation}
\begin{equation}
    f_{dot-att}(Q_i, K_i, V_i)=softmax(\frac{(Q_i^T K_i)}{\sqrt{d}})V_i
    \label{equation:eq3}
\end{equation}
\begin{equation}
    softmax(Z_i)=\frac{e^{Z_i}}{(\sum_{j=1}^d e^{Z_j})} \qquad i=1, 2, …, P
    \label{equation:eq4}
\end{equation}

\begin{figure}
  \centering
  \includegraphics{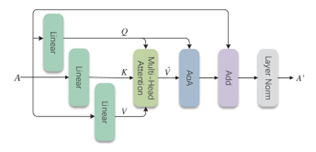}
  \caption{The AoANet refining module [20].}
  \label{fig:fig3}
\end{figure}

\begin{figure}
  \centering
  \includegraphics{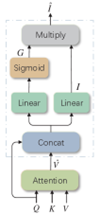}
  \caption{Attention on attention block details [20].}
  \label{fig:fig4}
\end{figure}

Considering $f_{mh-att}$ to denote the multi-head attention function. $Q,K,V \in R^{N \times M}$ are the linear projections of the input features, where $N$ is the number of feature vectors and M is the dimension of each feature vector. Concat is the concatenation function. $H$ is the number of attention heads. $Q_i$, $K_i$, $V_i$ are the linear transformations of $Q, K, V$ in each head.\\
You et al. [26] extracted global features and a list of visual attributes from an input image using deep CNN and a set of visual attribute detectors and used them in their proposed semantic attention model for image captioning. Anderson et al. [19] proposed a bottom-up and top-down attention model for image captioning. The bottom-up mechanism is applied to image regions, while the top-down mechanism calculates feature weights. Huang et al. [20] proposed the attention on attention (AoA) block to overcome the problems of the attention mechanism where there is no valuable information in the candidate vectors for attention block and it misled the decoder and results in irrelevant captions. They used AoA block in their proposed AoANet which is also based on the encoder-decoder approach. The encoder first extracts the image features (bottom-up features). Instead of feeding these vectors directly to the decoder, they proposed a refining network containing an AoA block to refine the input features representation. If $A$ shows the image’s feature vectors, the refining module output, $A^\prime$, is calculated using Equation \ref{equation:eq5}. The refining module in the encoder has been illustrated in Figure \ref{fig:fig3}. The components constituting AoA block are shown in Figure \ref{fig:fig4}.

\begin{equation}
    A^\prime=LayerNorm(A+AoA^E(f_{mh-att} (W^{Q_e} A, W^{K_e} A, W^{V_e} A)))
    \label{equation:eq5}
\end{equation}


In the above equation $W^{Q_e}$, $W^{K_e}$ and $W^{V_e}$ indicate three weight metrices to obtain linear projections of the feature vectors A, and $AoA^E$ stands for AoA-based encoding (see Figure \ref{fig:fig3}). The $LayerNorm$ operation normalizes its input. The multi-head attention module tries to calculate the interaction among objects in the refining module, and AoA block estimates how well they are related [20]. Then, the refined feature vectors are fed into the decoder to generate the final caption.\\
The AoA block details are shown in Figure \ref{fig:fig4}. The AoA module generates “information vectors” I and “attention gates” G. Each vector in these two sets is calculated by two different linear transformations on the attention result and the query q (Equations \ref{equation:eq6} and \ref{equation:eq7}). In these equations, $W_q^i,W_v^i,W_q^g,W_v^g \in R^{d \times d}$ are learnable attention weights, and $b^i,b^g \in R^d$ are the attention biases, where d is the dimension of $q$ and $v$ vectors, and $\hat{v}$ is the results of the attention block. The final AoA block output is calculated using Equation \ref{equation:eq8}, where $\cdot$ indicates element-wise multiplication.

\begin{equation}
    i=W_q^i q+W_v^i\hat{v}+b^i
    \label{equation:eq6}
\end{equation}
\begin{equation}
    g=\sigma(W_q^g q+W_v^g\hat{v}+b^g)
    \label{equation:eq7}
\end{equation}
\begin{equation}
    \hat{I}=G \cdot I
    \label{equation:eq8}
\end{equation}

\subsection{Visual Relationships and Scene Graphs}
Although the image objects are essential for the image captioning process, the object’s relations can enrich the generated captions. Thus, detecting these relations and using them in the captioning process would be important. One of the best ways to represent the visual relationships is using scene graphs. As shown in Figure \ref{fig:fig1}, the scene graph nodes are image entities, and their edges are the relationship between two entities. Sometimes the leaves are the entities attributes. To extract visual relationships, Lu et al. [55] trained a visual representation module and a language module and then combined them to predict the visual relationships. Yu et al. [56] obtained linguistic knowledge from training annotations and public text and used that to learn a visual relationship detection model. Liang et al. [57] proposed deep variation-structured reinforcement learning to capture global interdependency. Tang et al. [30] proposed an unbiased scene graph generator to tackle the biased relationships that exist in most scene graph generator models. For instance, the relationships like “on”, “has” and “near” are most likely to appear in the results of the inference phase instead of relations like “in front of” or “parked on”.\\
Using the detected visual relationships in the captioning process is challenging, and many approaches tried to address this problem. Chen et al. [58] employed an abstract scene graph (ASG) for image captioning. They introduced ASG2Caption model based on an encoder-decoder framework to generate captions directly from the ASG. ASG is a scene graph which its nodes do not have any label. Wang et al. [53] proposed a graph neural network-based model to implicitly model the visual relationships. First, they extract image features using a deep CNN and then consider each region of interest as a node and then build a relationship graph. Li et al. [21] use scene graphs to extract the visual relationship triplets. These information and visual object features extracted by Faster R-CNN are fed into a hierarchical attention model to generate the final captions. This method uses bottom-up features as spatial features; Thus, the drawbacks of these features still exist. The hierarchical attention model used in this method is based on simple attention blocks and as a result it suffers from the attention problems that are mentioned earlier.

\subsection{Deep Reinforcement Learning for Image Captioning}
The goal of a reinforcement learning agent is to maximize the total expected reward in an environment. Deep reinforcement learning can be used to optimize the captioning model by addressing the exposure bias problem and directly optimizing the non-differentiable image captioning evaluation metrics. Shi et al. [15] used policy and value networks in deep reinforcement learning for image captioning and optimized them using the temporal-difference approach. They trained the policy network using cross-entropy and value network using mean squared loss. Ren et al. [16] presented a decision-making framework for image captioning. They used an actor-critic algorithm to train the deep reinforcement learning agent using policy and value networks. An image-caption embedding is used as a reward function. Their policy and value network consist of a CNN followed by an RNN. Rennie et al. [18] introduced a REINFORCE-based self-critical sequence training (SCST) algorithm. The goal of training using REINFORCE algorithm is to minimize the negative expected reward. In practice $L(\theta)$ is estimated using a single sample from policy $p_\theta$ where $\theta$ is the parameter of the policy network. Equation \ref{equation:eq9} shows the calculation of $L(\theta)$ where $\omega^s=(\omega_1^s,…,\omega_T^s)$, and $r(\omega^s)$ is the reward of $\omega^s$. The REINFORCE algorithm is using a baseline to reduce the variance of the gradient estimate. Using the REINFORCE with baseline, the estimate of the gradient is calculated using Equation \ref{equation:eq10}. Where b is the baseline and can be any function which does not depend on the action. $h_t$ is the output of LSTM hidden unit in time step t. The symbol $1_{\omega_t^s}$ denotes the greatest probability (argmax) of $\omega_t^s$, the output of softmax and the network prediction for image caption tokens in time step t [18]. Rennie et al. [18] used a test-time inference reward threshold to normalize the rewards rather than estimating how the reward signal should be normalized, i.e. SCST uses current model test-time inference reward as the baseline in REINFORCE algorithm.

\begin{equation}
    L(\theta)= -r(\omega^s), \omega^s \sim p_\theta
    \label{equation:eq9}
\end{equation}
\begin{equation}
    \frac{\partial L(\theta)}{\partial s_t}=(r(\omega^s)-b)(p_\theta(\omega_t| h_t)-1_{\omega_t^s})
    \label{equation:eq10}
\end{equation}

\section{THE PROPOSED APPROACH}
Extraction of computationally intensive bottom-up features provides the captioning system with a large amount of information within the feature-maps obtained for many objects in the image. Processing such huge information will place a high computational burden on the captioning system. On top of that, from the final captioning point of view many of these features may be irrelevant or redundant. Moreover, bottom-up features do not encode the relationships between the objects in the image, which as explained earlier, are very important in generating correct captions. Moreover, due to the high computational cost of extracting these features, using them in an end-to-end manner would be challenging and thus such features are computed beforehand. As a result, their computation and selection cannot be directly incorporated in the training phase of the captioning system. \\
The AoANet architecture [20] has achieved remarkable results in image captioning. Introducing AoA block and utilizing this module in encoding and decoding process gives the model the ability of better refining the input features and generate better results. However, using hard-to-extract bottom-up features without considering the visual relationships is a significant drawback of this system. To address this problem, in this paper we introduce a scene-graph-based architecture based on encoder-decoderwhich takes advantage from the AoA module. The input features are the spatial and visual relationship features. The features are the whole image feature maps extracted by feeding the images into a pre-trained convolutional neural network. The visual relationship features are expressed in the form of triplets, which in the following sub-section we will discuss how to extract and prepare them for the network. We also introduced a multi-modal reward function for self-critical sequence training to achieve a more effective optimization of the model using deep reinforcement learning.

\subsection{Visual Relationship Features}
\begin{figure}
  \centering
  \includegraphics{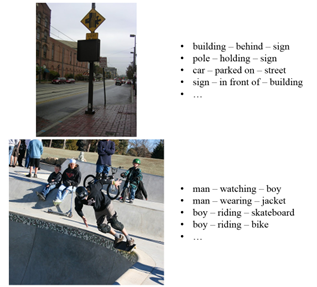}
  \caption{Examples of the relationship triplets generated for two sample images.}
  \label{fig:fig5}
\end{figure}

Visual relationships play an important role in image captioning. Mentioning visual relationships in the generated captions makes them rich from a semantic point of view. That's why considering them as input for the deep learning models provides more informative features and results in better captions. \\
To extract visual relationship features, a scene graph generator [30], [31], [59] is used to extract the word triplets (e.g., “man riding car”) where the first and the last words are the image objects, and the middle word is the relationship between these two objects. Figure \ref{fig:fig5} shows some of the relationship triplets extracted for two sample images. For each of these triplets a confidence score is computed and the top relationships based on their confidence scores are selected for each image. So, each image has a set of triplets as its visual relationship features, and each triplet has 3 words. \\
After extracting the triplets, they should be represented as numerical vectors before feeding them into the network. One of the best ways to represent each word as a vector is word embedding methods like Glove [60]. After embedding the words, each triplet is represented by three separate vectors, and aggregating them results in a single vector representing the whole relationship. Thus, the final visual relationship features can be concatenated to form a feature matrix.

\subsection{Scene-graph-based Attention on Attention Encoder}
\begin{figure}
  \centering
  \includegraphics{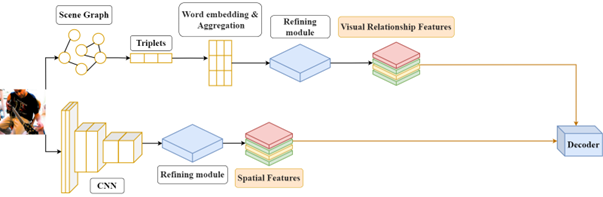}
  \caption{Our proposed network encoder to process both spatial and visual relationship features.}
  \label{fig:fig6}
\end{figure}

The details of the proposed network is explained here and the following sections. The inputs of this model are both image spatial low-level features extracted by a CNN and the visual relationship features introduced earlier. The visual relationships contain high-level semantic information of the image and help the model to achieve more reliable and improved results. \\
Our model’s encoder contains two paths to refine each feature set separately: One direction for spatial features, and the other for the visual relationship features. Each path includes a refining module that uses attention and AoA blocks to refine the input features. The diagram of this encoder is illustrated in Figure \ref{fig:fig6}. The top path refines the spatial features, whereas the bottom path is in charge of refining the visual relationship features. The refining module takes advantages of AoA blocks to obtain more semantic and useful features from the image. The refined features are then fed into the decoder. The decoder is in charge of generating the final caption. We have two separated matrices for both spatial and visual relationship features. An aggregation step is then applied on these feature maps to reduce their dimensions to vectors of equal size. These vectors are used in initializing the first LSTM cell in the decoder, as explained in the next sub-section.

\subsection{Scene-graph-based Attention on Attention Decoder}
The decoder takes the refined spatial and visual relationship features and generates the caption sequence $\omega$. The conditional probabilities of vocabulary entries in time step t are computed using Equation \ref{equation:eq11}.
\begin{equation}
    p(\omega_t|\omega_{1:t-1}, I)=softmax(W_pc_t)
    \label{equation:eq11}
\end{equation}

\begin{figure}
  \centering
  \includegraphics{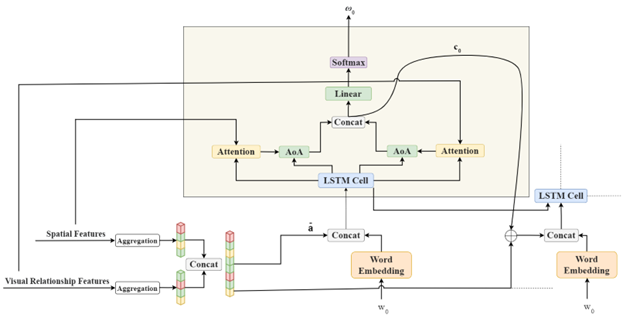}
  \caption{The proposed network decoder}
  \label{fig:fig7}
\end{figure}

Where $\omega_{1:t-1}$ are the generated tokens until time step $t-1$, and $I$ is the image features. $W_p$ is the learnable weight parameter, and $c_t$ is a context vector consisting of the decoding state and the new information acquired in each time step $t$ from both spatial and visual relationship features. \\
The decoder of our network is illustrated in Figure \ref{fig:fig7}. A visual vector initializer ($\bar{a}$) is first computed by concatenating the spatial features and visual relationship features aggregated vectors and fed into the model to initialize the first LSTM in time step $t_0$. This vector contains both spatial and visual relationship features and can be a good choice for initialization to input the visual information to the decoder network. It is vital to consider these information in all time steps as these information represent both spatial and visual relationship features, and feeding them into the network in each time step prevents information from forgetting them. Therefore, $\bar{a}$ is used in all time steps by adding this vector to the previous time steps context vector $c_{t-1}$. The resulting vector is called visual vector. \\
The inputs of LSTM in each time step consist of the word embedding of the current time step token ($\omega_t$) taken from ground truth caption concatenated with the visual vector $Concat(\bar{a}, c_{t-1})$ and the previous values of LSTM cell and hidden state. A trainable word embedding cell is used in to embed the captions tokens. The result of the word embedding cell is then concatenated to the visual vector and fed into the LSTM cell. We have two separated attention and AoA blocks to process both spatial and visual relationship features and use their information in the decoding process. The hidden state of the LSTM block is used as Q in both attention blocks. Then the spatial and visual relationship features are used as K and V in their decoding path's attention blocks. In this way, we calculate the similarity between K and Q and then apply the weights to V. To overcome the aforementioned attention drawback, attention on attention blocks are used on top of the attention blocks in decoding process. It is important to know that there are two separated decoding paths for both spatial and visual relationship features. The results of AoA blocks are then concatenated to each other. The resulting vector is the context vector $c_t$ of the current time step t. This vector is used to generate that time step's token and also fed into the next time step to compute the visual vector.

\subsection{Self-Critical Sequence Training with a Multi-Modal Reward Function}
\begin{figure}
  \centering
  \includegraphics{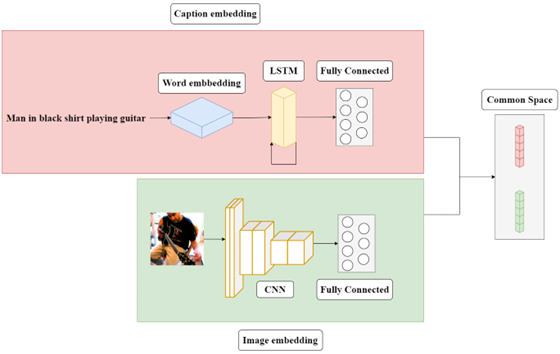}
  \caption{Visual semantic embedding network}
  \label{fig:fig8}
\end{figure}

Self-critical sequence training baselines the REINFORCE algorithm with the reward obtained by the current model at test time. The SCST has all the advantages of REINFORCE algorithm, but it does not need to learn an estimation of future rewards for baseline. The base SCST is used CIDEr as reward. CIDEr compares the generated caption by the ground truth captions. If the model generates a good caption which is not similar to ground truth captions, the CIDEr gives a low reward to this caption. To overcome this problem, we propose a multi-modal reward function for the SCST algorithm. In this work, instead of optimizing only CIDEr and considering it as the reward for the caption generating process, a linear combination of language and vision rewards are used as the final system reward and the RL agent goal is to maximize this reward. The vision reward computes the similarity between the generated caption and the image. Thus, it helps the RL agent to generate more meaningful and related captions. CIDEr is considered as language reward. Inspired by [16], an image-caption embedding network is used to embed images and captions into a common space and compare them using a similarity metric. The image-caption embedding network is illustrated in Figure \ref{fig:fig8}. The higher the similarity between the embedded caption and image is, the better the caption.

A pre-trained CNN is used to extract image features and to embed them in a common vector space. Also, we used the last cell output of an LSTM network as the representation of the caption, and embed it into the common vector space as well. In a training batch, there are different image-caption pairs. The goal is to maximize the similarity between the image-caption pairs and minimize the similarity between an image and other captions (and a caption and other images) in the batch. The loss function to train image-caption embedding network is presented in Equation \ref{equation:eq12}, where $\delta$ is the network parameters, $\bar{\omega}$ is a negative caption for the image $I$, the image $\bar{I}$ is a negative image for the caption \ref{equation:eq13}, and $i_e$ and $\omega_e$ are the embedding results of $I$ and $\omega$ respectively [16]. $\beta$ is called cross-validate margin and is used to generalize the embedding network. To calculate vision reward $r_v$, vectors are compared using a similarity metric (e.g., cosine similarity shown in Equation \ref{equation:eq14}).
The final reward is computed using Equation \ref{equation:eq14}. $r_l$ is language reward and can be an evaluation metric (e.g., CIDEr). The hyperparameter $\alpha$, controls the impact of language and vision components on the final reward, that when it is 1 the algorithm is like the basic SCST, and when it is 0, the algorithm only uses vision reward. In all equations $\cdot$ is the dot product.

\begin{equation}
    J(\delta)=\sum_I\sum_{\omega^-}max(0, \beta-I_e \cdot \omega_e+I_e \cdot \omega_e^-)+\sum_{\omega}\sum_{I^-}max(0, \beta - \omega_e \cdot I_e+\omega_e \cdot I_e^-)
    \label{equation:eq12}
\end{equation}
\begin{equation}
    r_v(\omega, i)=similarity(\omega_e, i_e )=\frac{\omega_e \cdot i_e}{\| \omega_e\| \|i_e\|}
    \label{equation:eq13}
\end{equation}
\begin{equation}
    r(\omega, I)= \alpha \cdot r_l(\omega)+(1-\alpha) \cdot r_v (\omega, I)
    \label{equation:eq14}
\end{equation}

Based on the REINFORCE algorithm, the gradient of the negative reward of a sample $\omega^s$ w.r.t. the softmax activations $s_t$ at time step $t$ can be calculated using Equation \ref{equation:eq15}, where $I$ is the image features, $\hat{\omega}$ is the sample obtained under test-time inference of the policy $p_\theta$ where the model chooses each token in a greedy manner Using SCST, the occurrence probability of the samples from the model which have higher reward than $\hat{\omega}$ will be increased and the probability of samples which have less reward than $\hat{\omega}$ will be decreased.

\begin{equation}
    \frac{\partial L(\theta)}{\partial s_t}=(r(\omega^s, I)-r(\hat{\omega}, I))(p_\theta(\omega_t|h_t )-1_{\omega_t^s })
    \label{equation:eq15}
\end{equation}

\section{EXPERIMENTAL RESULTS}
\subsection{Datasets}
he proposed image captioning model is evaluated on MSCOCO [36] dataset. This dataset contains 82,783 images for the training set, 40,504 samples for the validation set, and 40,775 images for the test set. We used another data split [61], in which 5000 images assigned to the validation set, 5000 images reserved for the test set, and the rest considered for the training set. Each image in the dataset has about five human-generated captions. For the sake of simplicity and to reduce the vocabulary size, we dropped the words that occur less than five times in the whole dataset.

\subsection{Evaluation Metrics}
The well-known evaluation metrics, BLEU [62], ROUGE [63], METEOR [64] and CIDEr [65] are used in the experiments to evaluate our proposed model compared to other captioning systems. BLEU is a set of evaluation metrics that first designed for machine translation, but widely used in image captioning. It evaluates how close is the generated caption to the ground truth caption based on n-grams. Depending on whether unigram, bigram, trigram, or 4-gram are used, BLEU-1, BLEU-2, BLEU-3, or BLEU-4 is computed respectively. ROUGE is an evaluation metric originally designed for automatic summary generation evaluation. ROUGE matches the sequence of words and their n-grams with ground truth captions. METEOR is another evaluation metric which is again originally designed for machine translation task. CIDEr is an evaluation metric designed specifically for image captioning. It uses TF-IDF (Term Frequency – Inverse Document Frequency) to score the important terms and then compare the generated captions to ground truth captions.

\subsection{Training and Objectives}
The primary loss function used for supervised training of the proposed captioning model is cross-entropy (XE) loss, shown in Equation \ref{equation:eq16} where $\omega^t$ is the target ground truth word of the caption.

\begin{equation}
    Loss=-\sum_{i=1}^T \omega^t \cdot log(p_\theta(\omega_t^s|\omega_{1:t-1}^s))
    \label{equation:eq16}
\end{equation}

In the second phase SCST is used to optimize the model. We evaluate SCST with both CIDEr and the proposed multi-modal reward (MMR) separately to check the effectiveness of using MMR to optimize the image captioning model.

\subsection{Implementation Details}
The CNN for spatial feature extraction from the images is the well-known Resnet-101 [23] model, pre-trained on ImageNet [66], employed by removing the fully-connected classification layer and used the last layer feature maps. The hidden state dimension of LSTM in the decoder is set to 512, and the feature vectors are also mapped into a 512-D space in the aggregation step of the encoder. The dimension of the Glove embedding vectors is 300. As a result the dimension of our visual relationships features is $20 \times 300$ as we used the top 20 relationships extracted by an unbiased scene-graph generator [30]. For the sake of easiness and speed, the visual relationship features are pre-extracted. In this process, the model is first trained using cross-entropy loss and the deep reinforcement learning approach is used to train the model. The number of heads for multi-head attention blocks are set to H=8. \\
We trained our networkusing two-phase training process. In the first phase cross-entropy loss is used for a maximum of 50 epochs. An early stopping method is also applied where the training process is stopped if the CIDEr metric does not improve for 5 consecutive epochs. In this phase, the learning rate starts at $5 \times 10^{-4}$, and is decreased every 5 epochs. The batch size is 128. For the second phase of training (when the SCST with MMR is used), the batch size is set to 64 and the learning rate is initialized to $2 \times 10^{-5}$. In this phase the model is trained for 30 epochs with early stopping. \\
The vision reward is calculated using cosine similarity. The image-caption embedding network, embed both images and captions into a common 512-D vector space. The networks hyperparameters have been chosen according to our limited training resources. However, all of the results have been achieved in a completely fair environment. The network hyperparameters are all the same in all of the experiments. The best value for $\alpha$ in Equation \ref{equation:eq14} is 0.7 according to our tests. Comparing to other papers, our hidden layers are chosen based on our limited computation resources. For instance, all of our hidden layers and common spaces are 512-dimension spaces, however in these dimensions in AoANet paper are 1024. This difference in dimension results in lower variance of the model.

\subsection{Scene-graph-based Attention on Attention Results}
\begin{figure}
  \centering
  \includegraphics{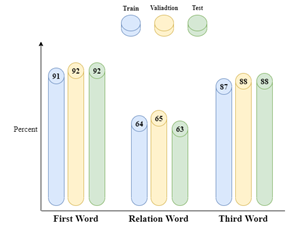}
  \caption{Occurrence of triplets words in the training, validation, and test sets.}
  \label{fig:fig9}
\end{figure}

Although the spatial features are essential for image captioning, the results show that those are not informative enough. Using the extracted triplets as visual relationship features along with the Resnet-101 extracted feature maps improve the model results. Thus, using extracted triplets from the scene graph makes the input features more informative. To have a better understanding of the information contained in the extracted triplets, compared with the ground truth captions, we checked the occurrence of each word of the triplets in the ground truth captions of MSCOCO dataset. \\
Figure \ref{fig:fig9} shows each triplets word occurrence in the training, validation, and test sets. These numbers show that the extracted relations from the scene graph are informative based on the human-generated captions. Thus, using them as visual relationship features are helpful.

Table \ref{tab:table1} shows that using the scene-graph-based attention on attention network improves the results compared to the base attention on attention model when it uses the same spatial features as the input. \\
As shown in Table \ref{tab:table2}, compared to the other models which use the CNN-based spatial feature maps, our model achieved more promising results. We also used LSTM instead of mean pooling to aggregate the triplets and generate a single vector from visual relationship triplets. The mean pooling approach shows better results. \\
Table \ref{tab:table2} also shows that our method achieved promising results compared to other models that use the spatial features as input. The results show the importance of using visual relationships for an image captioning system. The way we used the visual relationships does not deal with expensive graph calculations which is one of the strengths of our work. The results of the proposed network are even better than the AoA (using spatial features as input) when it optimized using SCST and multi-modal reward SCST. We mentioned several state-of-the-art models which are using CNN-based feature maps as the input features to compare with our proposed model. The results in Table \ref{tab:table2} show that our proposed model results are competitive to other models. CIDEr score, the most important metric for image captioning obtained by proposed methods, is significantly higher than other models. This means that using the visual relationship triplets is informative and helping the model to generate better captions. Although the bottom-up features based AoA results are a little bit better than ours, the ease of extracting and using the spatial and semantic features in the proposed network is another strength of our model, especially since is used in an end-to-end manner.

\begin{table}
 \caption{Our model results on MSCOCO “Karpathy” test split. The super-script SCST means the model optimized using Self-Critical Sequence Training in the two-phase training process and MMR stands for our proposed multi-modal reward. The models which use bottom-up features as input is marked using subscript Bottom-up. All the models (without substcript Bottom-up) are using CNN (e.g., Resnet-101, VGG-16) spatial features as input.}
  \centering
  \begin{tabular}{llllllll}
    Model    & B-1     & B-2     & B-3     & B-4     & METEOR     & ROUGE     & CIDEr \\
    \toprule
    $AoANet_{Bottom-up}$ & 74.1 & 57.4 & 42.9 & 31.7 & 25.3 & 54.3 & 102.3 \\
    \midrule
    $AoANet^{SCST}_{Bottom-up}$ & 76.9 &	60.5 &	45.6 &	34 & 26.1 &	55.5 &	110.5 \\
    \midrule
    $AoANet^{SCST+MMR}_{Bottom-up}$ & 77 &	60.4 &	45.7 & 34 & 26.4 &	55.6  &	112.6 \\
    \midrule
    $AoANet$ & 70.9 &	53.7 &	39.3 &	28.7 &	24.3 &	52 & 90.8 \\
    \midrule
    $AoANet^{SCST}$ & 72.2 &	55 &	40.4 &	29.7 &	24 & 52.6 &	94.8 \\
    \midrule
    $Ours^{LSTM}$ & 72.2 &	55 &	40.5 &	29.7 &	25.1 &	53 &	96.3 \\
    \midrule
    $Ours^{LSTM(SCST + MMR)}$ & 74.2	& 56.9	& 42.1 &	31 &	25.2 &	53.7 &	101.6 \\
    \midrule
    $Ours$ & 72.7  &	55.6  &	41 &	30.1 &	25.3 &	53.3 &	97.6 \\
    \midrule
    $Ours^{SCST}$ & 74.5 &	57.6 &	43.1 &	32.1 &	25.4 &	54.4 &	103.1 \\
    \midrule
    $Ours^{SCST+MMR}$ & 74.5 &	57.7 &	43	& 31.9 &	25.3 &	54.3 &	104.2 \\
    \bottomrule
  \end{tabular}
  \label{tab:table1}
\end{table}

\begin{table}
 \caption{Our models best results and other models' best results on MSCOCO “Karpathy” test sets.}
  \centering
  \begin{tabular}{llllllll}
    Model    & B-1     & B-2     & B-3     & B-4     & METEOR     & ROUGE     & CIDEr \\
    \toprule
    $NIC$ [6] & - &	- &	- &	27.7 &	23.7 &	- &	85.5 \\
    \midrule
    $gLSTM$ [28] & 67 &	49.1 &	35.8 &	26.4 &	22.7 &	- &	81.3 \\
    \midrule
    $Soft-Attention$ [7] & 70.7 &	48.9	& 34.4 &	24.3 &	23.9 &	- &	- \\
    \midrule
    $Hard-Attention$ [7] & 71.8 &	50.4	& 35.7 &	25 &	23.4 &	- &	- \\
    \midrule
    $AICRL-VGG16$ [49] & 70.2 &	53.6 &	39.8 &	29.5 &	23.6 &	- &	85.7 \\
    \midrule
    $AICRL-ResNet50$ [49] & 73.1 &	56.2 &	41 &	32.6 &	26.1 &	- &	87.2 \\
    \midrule
    $DRL$ [16] & 71.3 &	53.9 &	40.3	& 30.4 &	25.1 &	52.5 &	93.7 \\
    \midrule
    $Know More Say Less^{SCST}_{Bottom-up}$ [21] & 77.9 &	58.3 &	45.1 &	33.2	& 26.7 &	55.8 &	113.3 \\
    \midrule
    $Bottom-up Top-down^{SCST}_{Bottom-up}$ [19] & 79.8 &	- &	- &	36.3 &	27.7 &	56.9 &	120.1 \\
    \midrule
    $OFA$ [67] & -  &	- &	- &	41.4 &	30.8 &	- &	140.7 \\
    \midrule
    $mPLUG$ [68] & - &	- &	- &	46.5	& 32.0 &	- &	155.1 \\
    \midrule
    $AoANet*Bottom-up$ [20] & 76.9 &	60.5 &	42.6 &	34 &	55.5 &	26.1 &	110.5 \\
    \midrule
    $Ours^{SCST + MMR}$ & 74.5 &	57.7 &	43 &	31.9 &	25.3 &	54.3 &	104.2 \\
    \midrule
    $AoANet^{SCST + MMR}_{Bottom-up}$ & 77  &	60.4  &	45.7  &	34	& 26.4  &	55.6 &	112.6 \\
    \bottomrule
  \end{tabular} \\
 * Due to our resources limitations and for a fair comparison, we evaluated the AoANet in a completely similar environment and hyperparameters to our proposed network (e.g., Fully-connected nodes).
  \label{tab:table2}
\end{table}

\subsection{Multi-Modal Reward for Self-Critical Sequence Training}
In the two-phase training process, after training using cross-entropy loss, the model is optimized using DRL, specifically SCST. We used the basic SCST and also our proposed multi-modal reward function. The results show that the multi-modal reward-based SCST achieved better results in almost all the models. We have also trained the AoANet using bottom-up features under the same conditions as our networkwas trained. We optimized this model using both basic SCST and multi-modal reward-based SCST. It can be seen that using multi-modal reward-based SCST with the bottom-up features based AoANet achieved better results than basic SCST.

\begin{figure}
  \centering
  \includegraphics{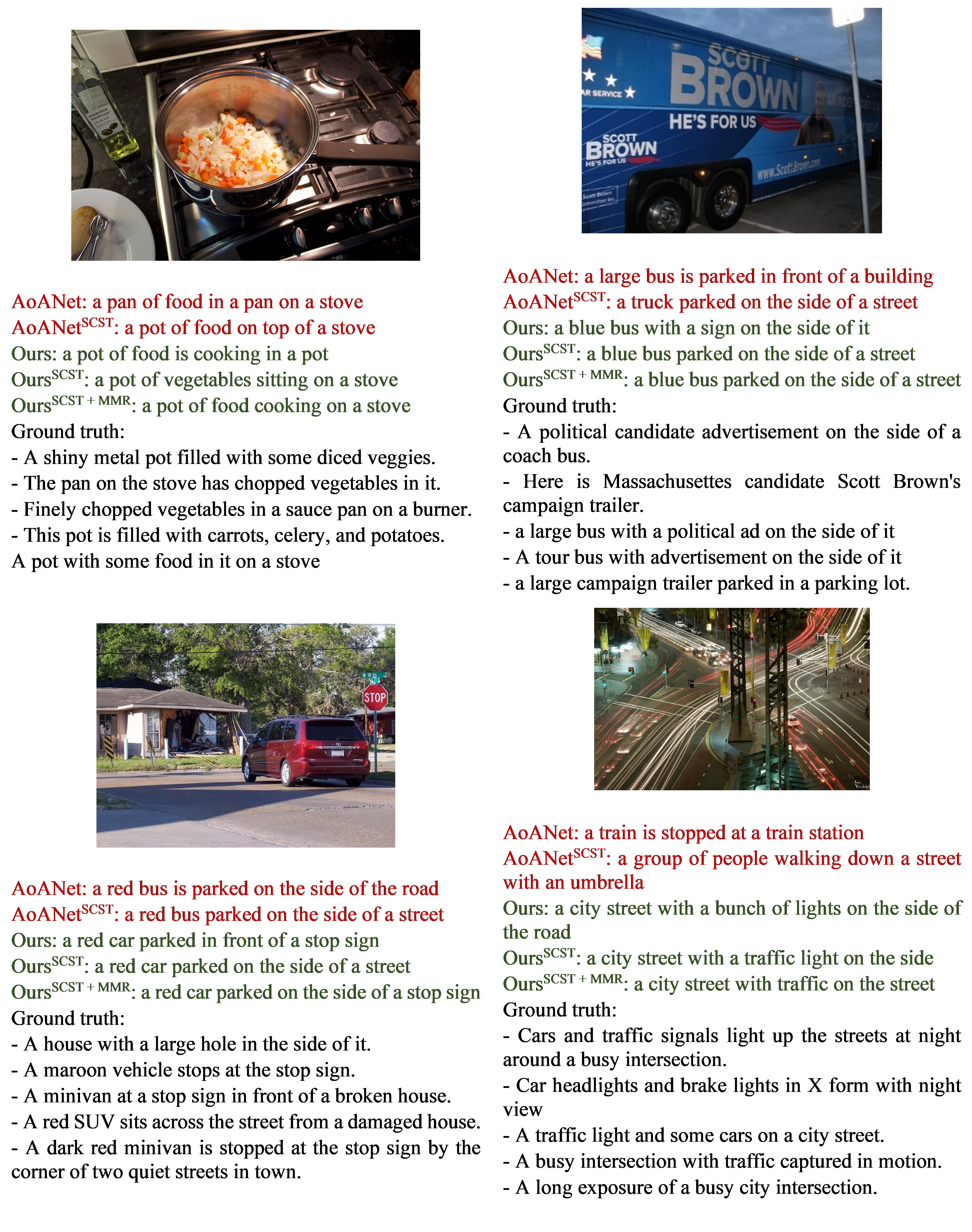}
  \caption{The generated captions by different method for several sample images.}
  \label{fig:fig10}
\end{figure}

\subsection{Qualitative Results}
To show the effect of using visual relationships features in the proposed network and also the effectiveness of multi-modal reward SCST, several captioning results generated by the proposed models have been demonstrated in Figure \ref{fig:fig10}. In this figure, the $AoANet$ is the same network as the network proposed by Huang et al. [20], but the input features are image spatial features instead of bottom-up features. We compare our proposed network with this model to show the effectiveness of using visual relationship features. Using visual relationship features, the image captioning network generates more meaningful captions and identifies more relations, enriching the captioning process. The visual relationship features improved the captions from a semantic point of view. They helped the image captioning system generate the captions more correctly as they have been used in caption generation process directly in our proposed network. From the results, it can be deduced that these features are informative, and using them help the captioning process. Also, optimizing using multi-modal SCST makes the generated captions more informative and improves the results. \\
For example, in the first column of the first row in Figure \ref{fig:fig10}, the base system mentioned “pan” twice, affecting the caption's meaning. When using the $AoANet$ model that has been optimized using SCST, the caption becomes correct from the meaning point of view, but the relationships are still not informative. When using ours, the caption mentions “cooking” which is very informative and salient in the image. When we optimize the model using SCST, although the “cooking” disappears from the caption, the caption mentioned “vegetables" which is essential. Moreover, the relationship between “pot” and “stove” has been explained, enriching the caption. The best caption is generated using the proposed network model after optimizing using multi-modal reward SCST. The generated caption mentioned “cooking” and also mentioned the relation between “pot” and “stove”. \\
In the second column on the first row in Figure \ref{fig:fig10}, the AoANet does not mention the bus color or its relationship with the sign or the street. Although the $AoANet^{SCST}$ mentioned the street, it said “truck” instead of the bus. When we used our proposed netwrok, not only the color of the bus is mentioned, but also its relationship with the sign is noted. $Ours^{SCST}$ model generates a caption that mentions the bus color and its relationship with the street. $Ours^{SCST + MMR}$ generates the caption same as $Ours^{SCST}$. The first row is images from the validation split of the MSCOCO dataset.
The second row contains images from the test set of MSCOCO dataset. As seen in the first column of the second row, the $AoANet$ and $AoANet^{SCST}$ models have mentioned a bus instead of a car in the caption. The SGrAtten mentioned the car correctly, and the relationship between the car and the “stop sign” which is “in front of” is mentioned as well. $Ours^{SCST}$ mentioned the street in the image, and the $Ours^{SCST + MMR}$ said “side of” as the relation between “car” and the “stop sign” which is more precise than “in front of”. In the second column of the second row, the base $AoANet$ model ones generated completely irrelevant captions, but our proposed models have generated completely meaningful captions with meaningful relationships. It is observed that the best caption generated by $Ours^{SCST + MMR}$ is semantically and grammatically correct. \\
As can be seen, our proposed network results are adequate, and the generated captions using visual relationship features are more comprehensive. Moreover, the multi-modal reward function improved the SCST, and as a result, the models are optimized better using this deep reinforcement learning-based algorithm. Generating visual relationship features is not very complicated, making them perfect features to use in end-to-end image captioning systems.

\section{CONCLUSION}
In this work, we proposed a scene-graph-based attention on attention image captioning method with a multi-modal reward function for self-critical sequence training algorithm. We introduced visual relationship features to improve the model results. These features are used along with spatial low-level features extracted by feeding the whole image into a pre-trained CNN network. Although most image captioning methods use bottom-up features, those are hard to use in end-to-end systems. We tried to use the whole image feature maps to address this problem, but the lack of semantic information in these features is a major drawback. Thus, a scene graph generator is proposed to be used to extract the images visual relationships and use the obtained relationships triplets as the visual relationship features along with the spatial features to guide the system to achieve more accurate captions. The results show that our proposed network generates more comprehensive captions than the baseline models, as well as the other models with CNN feature maps as input features. Also, the results show that the multi-modal reward SCST optimized the model better than SCST, and the results were utterly improved when this algorithm used to optimize the model. We also optimized the bottom-up feature based AoA using multi-modal reward SCST to show effectiveness of this reward. For the subsequent research, we intend to study extracting even more informative semantic attributes from the scene graph and check the possibility of using these features to guide the system to improve the models with bottom-up features.

\section*{References}
[1]	M. Z. Hossain, F. Sohel, M. F. Shiratuddin, and H. Laga, “A Comprehensive Survey of Deep Learning for Image Captioning,” ACM Comput. Surv., Oct. 2018, Accessed: May 29, 2021. [Online]. Available: http://arxiv.org/abs/1810.04020 \\\
[2]	H. Wang, Y. Zhang, and X. Yu, “An Overview of Image Caption Generation Methods,” Comput. Intell. Neurosci., vol. 2020, pp. 1–13, Jan. 2020, doi: 10.1155/2020/3062706. \\\
[3]	S. Iyer, S. Chaturvedi, and T. Dash, “Image Captioning-Based Image Search Engine: An Alternative to Retrieval by Metadata,” in Soft Computing for Problem Solving, vol. 817, J. C. Bansal, K. N. Das, A. Nagar, K. Deep, and A. K. Ojha, Eds. Singapore: Springer Singapore, 2019, pp. 181–191. doi: 10.1007/978-981-13-1595-4\textunderscore14. \\\
[4]	S. P. Manay, S. A. Yaligar, Y. Thathva Sri Sai Reddy, and N. J. Saunshimath, “Image Captioning for the Visually Impaired,” in Emerging Research in Computing, Information, Communication and Applications, vol. 789, N. R. Shetty, L. M. Patnaik, H. C. Nagaraj, P. N. Hamsavath, and N. Nalini, Eds. Singapore: Springer Singapore, 2022, pp. 511–522. doi: 10.1007/978-981-16-1338-8\textunderscore43. \\\
[5]	N. Vijayaraju, “Image Retrieval Using Image Captioning,” Master of Science, San Jose State University, San Jose, CA, USA, 2019. doi: 10.31979/etd.vm9n-39ed. \\\
[6]	O. Vinyals, A. Toshev, S. Bengio, and D. Erhan, “Show and Tell: A Neural Image Caption Generator,” in Proceedings of the IEEE conference on computer vision and pattern recognition, Apr. 2015. Accessed: May 29, 2021. [Online]. Available: http://arxiv.org/abs/1411.4555 \\\
[7]	K. Xu et al., “Show, Attend and Tell: Neural Image Caption Generation with Visual Attention,” in Proceedings of the 32nd International Conference on International Conference on Machine Learning - Volume 37, Apr. 2016. Accessed: May 30, 2021. [Online]. Available: http://arxiv.org/abs/1502.03044 \\\
[8]	Y. Su, Y. Li, N. Xu, and A.-A. Liu, “Hierarchical Deep Neural Network for Image Captioning,” Neural Process. Lett., vol. 52, no. 2, pp. 1057–1067, Oct. 2020, doi: 10.1007/s11063-019-09997-5.\\\
[9]	J. Donahue et al., “Long-term recurrent convolutional networks for visual recognition and description,” in 2015 IEEE Conference on Computer Vision and Pattern Recognition (CVPR), Boston, MA, USA, Jun. 2015, pp. 2625–2634. doi: 10.1109/CVPR.2015.7298878.\\\
[10]	N. Xu et al., “Multi-Level Policy and Reward-Based Deep Reinforcement Learning Framework for Image Captioning,” IEEE Trans. Multimed., vol. 22, no. 5, pp. 1372–1383, May 2020, doi: 10.1109/TMM.2019.2941820. \\\
[11]	K. Cho et al., “Learning Phrase Representations using RNN Encoder-Decoder for Statistical Machine Translation,” in 2014 Conference on Empirical Methods in Natural Language Processing (EMNLP), Sep. 2014. Accessed: May 29, 2021. [Online]. Available: http://arxiv.org/abs/1406.1078 \\\
[12]	Y. Pu et al., “Variational Autoencoder for Deep Learning of Images, Labels and Captions,” ArXiv160908976 Cs Stat, Sep. 2016, Accessed: May 29, 2021. [Online]. Available: http://arxiv.org/abs/1609.08976 \\\
[13]	X. Xiao, L. Wang, K. Ding, S. Xiang, and C. Pan, “Deep Hierarchical Encoder–Decoder Network for Image Captioning,” IEEE Trans. Multimed., vol. 21, no. 11, pp. 2942–2956, Nov. 2019, doi: 10.1109/TMM.2019.2915033. \\\
[14]	S. Bengio, O. Vinyals, N. Jaitly, and N. Shazeer, “Scheduled Sampling for Sequence Prediction with Recurrent Neural Networks,” ArXiv150603099 Cs, Sep. 2015, Accessed: Dec. 12, 2021. [Online]. Available: http://arxiv.org/abs/1506.03099 \\\
[15]	H. Shi, P. Li, B. Wang, and Z. Wang, “Image Captioning based on Deep Reinforcement Learning,” p. 5. \\\
[16]	Z. Ren, X. Wang, N. Zhang, X. Lv, and L.-J. Li, “Deep Reinforcement Learning-Based Image Captioning with Embedding Reward,” in 2017 IEEE Conference on Computer Vision and Pattern Recognition (CVPR), Honolulu, HI, Jul. 2017, pp. 1151–1159. doi: 10.1109/CVPR.2017.128. \\\
[17]	K. Arulkumaran, M. P. Deisenroth, M. Brundage, and A. A. Bharath, “Deep Reinforcement Learning: A Brief Survey,” IEEE Signal Process. Mag., vol. 34, no. 6, pp. 26–38, Nov. 2017, doi: 10.1109/MSP.2017.2743240. \\\
[18]	S. J. Rennie, E. Marcheret, Y. Mroueh, J. Ross, and V. Goel, “Self-critical Sequence Training for Image Captioning,” ArXiv161200563 Cs, Nov. 2017, Accessed: May 29, 2021. [Online]. Available: http://arxiv.org/abs/1612.00563 \\\
[19]	P. Anderson et al., “Bottom-Up and Top-Down Attention for Image Captioning and Visual Question Answering,” in 2018 IEEE/CVF Conference on Computer Vision and Pattern Recognition, Mar. 2018. Accessed: May 29, 2021. [Online].  Available: http://arxiv.org/abs/1707.07998 \\\
[20]	L. Huang, W. Wang, J. Chen, and X.-Y. Wei, “Attention on Attention for Image Captioning,” in 2019 IEEE/CVF International Conference on Computer Vision (ICCV), Seoul, Korea (South), Oct. 2019, pp. 4633–4642. doi: 10.1109/ICCV.2019.00473. \\\
[21]	X. Li and S. Jiang, “Know More Say Less: Image Captioning Based on Scene Graphs,” IEEE Trans. Multimed., vol. 21, no. 8, pp. 2117–2130, Aug. 2019, doi: 10.1109/TMM.2019.2896516. \\\
[22]	S. Ren, K. He, R. Girshick, and J. Sun, “Faster R-CNN: Towards Real-Time Object Detection with Region Proposal Networks,” presented at the Advances in Neural Information Processing Systems 28 (NIPS 2015), p. 14. \\\
[23]	K. He, X. Zhang, S. Ren, and J. Sun, “Deep Residual Learning for Image Recognition,” in IEEE Conference on Computer Vision and Pattern Recognition (CVPR), Dec. 2015. Accessed: May 29, 2021. [Online]. Available: http://arxiv.org/abs/1512.03385 \\\
[24]	Y. Bin, Y. Yang, J. Zhou, Z. Huang, and H. T. Shen, “Adaptively Attending to Visual Attributes and Linguistic Knowledge for Captioning,” in Proceedings of the 25th ACM international conference on Multimedia, Mountain View California USA, Oct. 2017, pp. 1345–1353. doi: 10.1145/3123266.3123391. \\\
[25]	F. Chen, R. Ji, J. Su, Y. Wu, and Y. Wu, “StructCap: Structured Semantic Embedding for Image Captioning,” in Proceedings of the 25th ACM international conference on Multimedia, Mountain View California USA, Oct. 2017, pp. 46–54. doi: 10.1145/3123266.3123275. \\\
[26]	Q. You, H. Jin, Z. Wang, C. Fang, and J. Luo, “Image Captioning with Semantic Attention,” ArXiv160303925 Cs, Mar. 2016, Accessed: Dec. 09, 2021. [Online]. Available: http://arxiv.org/abs/1603.03925 \\\
[27]	Q. Wu, C. Shen, L. Liu, A. Dick, and A. Van Den Hengel, “What Value Do Explicit High Level Concepts Have in Vision to Language Problems?,” in 2016 IEEE Conference on Computer Vision and Pattern Recognition (CVPR), Las Vegas, NV, Jun. 2016, pp. 203–212. doi: 10.1109/CVPR.2016.29. \\\
[28]	X. Jia, E. Gavves, B. Fernando, and T. Tuytelaars, “Guiding Long-Short Term Memory for Image Caption Generation,” ArXiv150904942 Cs, Sep. 2015, Accessed: Jun. 05, 2021. [Online]. Available: http://arxiv.org/abs/1509.04942 \\\
[29]	N. Xu, A.-A. Liu, J. Liu, W. Nie, and Y. Su, “Scene graph captioner: Image captioning based on structural visual representation,” J. Vis. Commun. Image Represent., vol. 58, pp. 477–485, Jan. 2019, doi: 10.1016/j.jvcir.2018.12.027. \\\
[30]	K. Tang, Y. Niu, J. Huang, J. Shi, and H. Zhang, “Unbiased Scene Graph Generation from Biased Training,” ArXiv200211949 Cs, Mar. 2020, Accessed: May 29, 2021. [Online]. Available: http://arxiv.org/abs/2002.11949 \\\
[31]	K. Tang, “A Scene Graph Generation Codebase in PyTorch.” 2020. \\\
[32]	N. Xu, “Scene Graph Captioner: Image Captioning Based on Structural Visual Representation,” p. 31. \\\
[33]	J. Johnson et al., “Image retrieval using scene graphs,” in 2015 IEEE Conference on Computer Vision and Pattern Recognition (CVPR), Boston, MA, USA, Jun. 2015, pp. 3668–3678. doi: 10.1109/CVPR.2015.7298990. \\\
[34]	Y. Jia, “Attention Mechanism in Machine Translation,” J. Phys. Conf. Ser., vol. 1314, p. 012186, Oct. 2019, doi: 10.1088/1742-6596/1314/1/012186. \\\
[35]	T. Luong, H. Pham, and C. D. Manning, “Effective Approaches to Attention-based Neural Machine Translation,” in Proceedings of the 2015 Conference on Empirical Methods in Natural Language Processing, Lisbon, Portugal, 2015, pp. 1412–1421. doi: 10.18653/v1/D15-1166. \\\
[36]	X. Chen et al., “Microsoft COCO Captions: Data Collection and Evaluation Server,” ArXiv150400325 Cs, Apr. 2015, Accessed: Oct. 08, 2021. [Online]. Available: http://arxiv.org/abs/1504.00325 \\\
[37]	G. Kulkarni et al., “Baby talk: Understanding and generating simple image descriptions,” in CVPR 2011, Colorado Springs, CO, USA, Jun. 2011, pp. 1601–1608. doi: 10.1109/CVPR.2011.5995466. \\\
[38]	S. Li, G. Kulkarni, T. L Berg, A. C. Berg, and Y. Choi, “Composing Simple Image Descriptions using Web-scale N-grams,” in Proceedings of the Fifteenth Conference on Computational Natural Language Learning, 2011, pp. 220–228. \\\
[39]	M. Mitchell et al., “Midge: Generating Image Descriptions From Computer Vision Detections,” in Proceedings of the 13th Conference of the European Chapter of the Association for Computational Linguistics, 2012. \\\
[40]	Y. Ushiku, T. Harada, and Y. Kuniyoshi, “Efficient image annotation for automatic sentence generation,” in Proceedings of the 20th ACM international conference on Multimedia - MM ’12, Nara, Japan, 2012, p. 549. doi: 10.1145/2393347.2393424. \\\
[41]	S. Bai and S. An, “A survey on automatic image caption generation,” Neurocomputing, vol. 311, pp. 291–304, Oct. 2018, doi: 10.1016/j.neucom.2018.05.080. \\\
[42]	J. Curran, S. Clark, and J. Bos, “Linguistically Motivated Large-Scale NLP with C\&C and Boxer,” in Proceedings of the 45th Annual Meeting of the Association for Computational Linguistics Companion Volume Proceedings of the Demo and Poster Sessions, 2007. \\\
[43]	R. Mason and E. Charniak, “Nonparametric Method for Data-driven Image Captioning,” in Proceedings of the 52nd Annual Meeting of the Association for Computational Linguistics (Volume 2: Short Papers), Baltimore, Maryland, 2014, pp. 592–598. doi: 10.3115/v1/P14-2097. \\\
[44]	M. Hodosh, P. Young, and J. Hockenmaier, “Framing Image Description as a Ranking Task Data, Models and Evaluation Metrics Extended Abstract,” p. 5. \\\
[45]	A. Gupta, Y. Verma, and C. V. Jawahar, “Choosing linguistics over vision to describe images,” presented at the Twenty-Sixth National Conference on Artificial Intelligence, 2012, pp. 606--612. \\\
[46]	J. Mao, W. Xu, Y. Yang, J. Wang, Z. Huang, and A. Yuille, “Deep Captioning with Multimodal Recurrent Neural Networks (m-RNN),” ArXiv14126632 Cs, Jun. 2015, Accessed: Dec. 12, 2021. [Online]. Available: http://arxiv.org/abs/1412.6632 \\\
[47]	X. Chen and C. L. Zitnick, “Mind’s eye: A recurrent visual representation for image caption generation,” in 2015 IEEE Conference on Computer Vision and Pattern Recognition (CVPR), Boston, MA, USA, Jun. 2015, pp. 2422–2431. doi: 10.1109/CVPR.2015.7298856.\\\
[48]	A. Vaswani et al., “Attention Is All You Need,” in 31st International Conference on Neural Information Processing Systems, Dec. 2017. Accessed: May 29, 2021. [Online]. Available: http://arxiv.org/abs/1706.03762\\\
[49]	Y. Chu, X. Yue, L. Yu, M. Sergei, and Z. Wang, “Automatic Image Captioning Based on ResNet50 and LSTM with Soft Attention,” Wirel. Commun. Mob. Comput., vol. 2020, pp. 1–7, Oct. 2020, doi: 10.1155/2020/8909458.\\\
[50]	L. Gao, X. Li, J. Song, and H. T. Shen, “Hierarchical LSTMs with Adaptive Attention for Visual Captioning,” IEEE Trans. Pattern Anal. Mach. Intell., pp. 1–1, 2019, doi: 10.1109/TPAMI.2019.2894139. \\\
[51]	J. Jin, K. Fu, R. Cui, F. Sha, and C. Zhang, “Aligning where to see and what to tell: image caption with region-based attention and scene factorization,” ArXiv150606272 Cs Stat, Jun. 2015, Accessed: Jun. 05, 2021. [Online]. Available: http://arxiv.org/abs/1506.06272 \\\
[52]	S. Wang, L. Lan, X. Zhang, G. Dong, and Z. Luo, “Object-aware semantics of attention for image captioning,” Multimed. Tools Appl., vol. 79, no. 3–4, pp. 2013–2030, Jan. 2020, doi: 10.1007/s11042-019-08209-5. \\\
[53]	J. Wang, W. Wang, L. Wang, Z. Wang, D. D. Feng, and T. Tan, “Learning visual relationship and context-aware attention for image captioning,” Pattern Recognit., vol. 98, p. 107075, Feb. 2020, doi: 10.1016/j.patcog.2019.107075. \\\
[54]	S. Yan, Y. Xie, F. Wu, J. S. Smith, W. Lu, and B. Zhang, “Image captioning via hierarchical attention mechanism and policy gradient optimization,” Signal Process., vol. 167, p. 107329, Feb. 2020, doi: 10.1016/j.sigpro.2019.107329. \\\
[55]	C. Lu, R. Krishna, M. Bernstein, and L. Fei-Fei, “Visual Relationship Detection with Language Priors,” ArXiv160800187 Cs, Jul. 2016, Accessed: Dec. 15, 2021. [Online]. Available: http://arxiv.org/abs/1608.00187 \\\
[56]	R. Yu, A. Li, V. I. Morariu, and L. S. Davis, “Visual Relationship Detection with Internal and External Linguistic Knowledge Distillation,” ArXiv170709423 Cs, Aug. 2017, Accessed: Dec. 15, 2021. [Online]. Available: http://arxiv.org/abs/1707.09423 \\\
[57]	X. Liang, L. Lee, and E. P. Xing, “Deep Variation-Structured Reinforcement Learning for Visual Relationship and Attribute Detection,” in Proceedings of the IEEE Conference on Computer Vision and Pattern Recognition (CVPR), Jul. 2017. \\\
[58]	S. Chen, Q. Jin, P. Wang, and Q. Wu, “Say As You Wish: Fine-grained Control of Image Caption Generation with Abstract Scene Graphs,” ArXiv200300387 Cs, Feb. 2020, Accessed: Jun. 11, 2021. [Online]. Available: http://arxiv.org/abs/2003.00387 \\\
[59]	K. Tang, H. Zhang, B. Wu, W. Luo, and W. Liu, “Learning to Compose Dynamic Tree Structures for Visual Contexts,” in Conference on Computer Vision and Pattern Recognition, 2019. \\\
[60]	J. Pennington, R. Socher, and C. D. Manning, “GloVe: Global Vectors for Word Representation,” in Empirical Methods in Natural Language Processing (EMNLP), 2014, pp. 1532–1543. [Online]. Available: http://www.aclweb.org/anthology/D14-1162 \\\
[61]	A. Karpathy and L. Fei-Fei, “Deep Visual-Semantic Alignments for Generating Image Descriptions,” ArXiv14122306 Cs, Apr. 2015, Accessed: Dec. 17, 2021. [Online]. Available: http://arxiv.org/abs/1412.2306 \\\
[62]	K. Papineni, S. Roukos, T. Ward, and W.-J. Zhu, “BLEU: a method for automatic evaluation of machine translation,” in Proceedings of the 40th Annual Meeting on Association for Computational Linguistics  - ACL ’02, Philadelphia, Pennsylvania, 2001, p. 311. doi: 10.3115/1073083.1073135. \\\
[63]	C.-Y. Lin, “ROUGE: A Package for Automatic Evaluation of Summaries,” in Text Summarization Branches Out, Barcelona, Spain, Jul. 2004, pp. 74–81. [Online]. Available: https://www.aclweb.org/anthology/W04-1013 \\\
[64]	A. Lavie and A. Agarwal, “Meteor: An Automatic Metric for MT Evaluation with High Levels of Correlation with Human Judgments,” in Proceedings of the Second Workshop on Statistical Machine Translation, USA, 2007, pp. 228–231. \\\
[65]	R. Vedantam, C. L. Zitnick, and D. Parikh, “CIDEr: Consensus-based image description evaluation,” in 2015 IEEE Conference on Computer Vision and Pattern Recognition (CVPR), 2015, pp. 4566–4575. doi: 10.1109/CVPR.2015.7299087. \\\
[66]	O. Russakovsky et al., “ImageNet Large Scale Visual Recognition Challenge,” ArXiv14090575 Cs, Jan. 2015, Accessed: May 31, 2021. [Online]. Available: http://arxiv.org/abs/1409.0575 \\\
[67]	P. Wang et al., “OFA: Unifying Architectures, Tasks, and Modalities Through a Simple Sequence-to-Sequence Learning Framework.” arXiv, Jun. 01, 2022. Accessed: Jan. 25, 2023. [Online]. Available: http://arxiv.org/abs/2202.03052 \\\
[68]	C. Li et al., “mPLUG: Effective and Efficient Vision-Language Learning by Cross-modal Skip-connections.” arXiv, May 25, 2022. Accessed: Jan. 25, 2023. [Online]. Available: http://arxiv.org/abs/2205.12005
\end{document}